\documentclass[sigconf]{acmart}

\usepackage{subcaption}
\usepackage{listings}
\AtBeginDocument{%
  \providecommand\BibTeX{{%
    \normalfont B\kern-0.5em{\scshape i\kern-0.25em b}\kern-0.8em\TeX}}}

\copyrightyear{2020} 
\acmYear{2020} 
\setcopyright{acmcopyright}\acmConference[FDG '20]{International Conference on the Foundations of Digital Games}{September 15--18, 2020}{Bugibba, Malta}
\acmBooktitle{International Conference on the Foundations of Digital Games (FDG '20), September 15--18, 2020, Bugibba, Malta}
\acmPrice{15.00}
\acmDOI{10.1145/3402942.3409606}
\acmISBN{978-1-4503-8807-8/20/09}

\begin{document}

%%
%% The "title" command has an optional parameter,
%% allowing the author to define a "short title" to be used in page headers.
\title{Multi-Objective level generator generation with Marahel}

%%
%% The "author" command and its associated commands are used to define
%% the authors and their affiliations.
%% Of note is the shared affiliation of the first two authors, and the
%% "authornote" and "authornotemark" commands
%% used to denote shared contribution to the research.
% \author{King Anonymous}
% \affiliation{%
%   \institution{The Royal Anonymous Academy}
%   \streetaddress{Anonymous St}
%   \city{Anonymous Capital}
%   \country{Anonymous Kingdom}
% }
% \email{anonymous@anonymouskingdom.com}

\author{Ahmed Khalifa}
\affiliation{%
  \institution{New York University}
  \streetaddress{370 Jay St}
  \city{Brooklyn}
  \state{New York}
  \postcode{11201}
}
\email{ahmed@akhalifa.com}

\author{Julian Togelius}
\affiliation{%
  \institution{New York University}
  \streetaddress{370 Jay St}
  \city{Brooklyn}
  \state{New York}
  \postcode{11201}
}
\email{julian@togelius.com}

%%
%% By default, the full list of authors will be used in the page
%% headers. Often, this list is too long, and will overlap
%% other information printed in the page headers. This command allows
%% the author to define a more concise list
%% of authors' names for this purpose.
\renewcommand{\shortauthors}{Khalifa and Togelius}
% \renewcommand{\shortauthors}{Anonymous, et al.}

%%
%% The abstract is a short summary of the work to be presented in the
%% article.
\begin{abstract}
This paper introduces a new system to design constructive level generators by searching the space of constructive level generators defined by Marahel language. We use NSGA-II, a multi-objective optimization algorithm, to search for generators for three different problems (Binary, Zelda, and Sokoban). We restrict the representation to a subset of Marahel language to push the evolution to find more efficient generators. The results show that the generated generators were able to achieve good performance on most of the fitness functions over these three problems. However, on Zelda and Sokoban they tend to depend on the initial state than modifying the map.
\end{abstract}

%%
%% The code below is generated by the tool at http://dl.acm.org/ccs.cfm.
%% Please copy and paste the code instead of the example below.
%%


%%
%% Keywords. The author(s) should pick words that accurately describe
%% the work being presented. Separate the keywords with commas.

%%
%% This command processes the author and affiliation and title
%% information and builds the first part of the formatted document.
\maketitle

\section{Introduction}

Designing good levels is hard, but designing good level generators is arguably harder. The requirements on a level generator vary, but in general it is expected to produce levels that not only meet certain quality criteria, but do it consistently and with a certain degree of diversity so as to not bore the player (or designer). Faced with such a design problem, the generatively minded thinker might consider solving it by creating a level generator generator~\cite{kerssemakers2012procedural}. 

In a search-based framework, this is not in principle much harder than creating a search-based generator. If one can formulate useful quality criteria for levels, these could be used for evolving a level generator itself. In other words, the fitness function for the generator measures the quality of its generated levels as a proxy for (or measure of) the quality of the generator.

The most obvious advantage of evolving a level generator, compared to simply evolving the individual levels, is generation speed: search-based PCG is quite slow, but an evolved generator can be much faster, in particular if it is a constructive generator. In this sense, time can be invested in evolving a generator and the investment later pays off when an arbitrarily large number of new levels can be generated in very little time. Another advantage is that finding a high-quality generator able to generate levels for a particular game can help us understand the design of the game itself, as it in some sense forms an abstraction of a space of good levels for the game. This, however, requires that the generator representation is such that a human can understand the evolved generator.

This paper describes a system for evolving level generators for 2D games. The system is based on \emph{Marahel}, a previously introduced language for constructive level generators. (The version used in this paper is somewhat expanded compared to the earlier published version.) Grammatical evolution, a form of genetic programming, is used to evolve Marahel programs, and these programs are then evaluated by letting them generate a number of levels and testing the levels. As there are multiple quality criteria, a multiobjective evolutionary algorithm is applied within the grammatical evolution framework. This system is applied to three different level generation problems: generating long paths and connected segments in a binary tilemap, generating levels for a simple version of the \emph{Legend of Zelda} dungeon system, and generating \emph{Sokoban} levels.

\section{Background}
Procedural content generation (PCG) is the process of creating a game content using a computer program with limited human input. PCG has been used for many different aspects of games such as textures~\cite{liang2001real,cohen2003wang}, rules~\cite{browne2010evolutionary,nielsen2015towards}, patterns~\cite{khalifa2018talakat,hastings2009evolving}, etc. PCG is usually divided based on the used methods. Each method has its own advantages and disadvantages. Three main types of PCG are: Constructive, Search-Based, and Machine Learning. Constructive approaches~\cite{shaker2016procedural} applies a set of rules to generate a content. These rules are designed by the game designer to follow and find a certain content. It is usually used in the game industry due to its generation speed and direct control on the generated content. Search-based approaches~\cite{togelius2011search} uses a search algorithm to find the required content. These approaches are guided using a fitness function which measures how close the current content to an ideal content. These approaches are mainly used in research and rarely in the industry due to the longer time needed to generate content and the indirect control of the generated content. Finally, Machine learning approaches~\cite{summerville2018procedural} uses machine learning techniques to generate the content. Similarly to search-based approaches, machine learning approaches are used more in research than in industry. This is due to the need for training time and training data.

\subsection{Procedural Procedural Level Generation Generation}
Procedural Procedural Level Generation Generation is the problem of using a procedural generation method to find a level generator. A key requirement of most procedural generators is a reasonable and workable representation of the generated content. For example: Browne and Maire~\cite{browne2010evolutionary} represented board games using Game Description Language (GDL) to be able to evolve new board games like Yavalath~\cite{brown2012yavalath}. Finding a representation for a level generator is arguably a harder problem as the representation should be able to encode many different types of generators that can produce different content. 

One of the early attempts to create procedural level generator generators~\cite{togelius2012compositional} searched the parameter space of ASP programs to generate dungeon crawler level generators that are challenging for an automated agent. Later, Kerssemakers et al.~\cite{kerssemakers2012procedural} designed a meta-generator for Super Mario Bros (Nintendo, 1985) and used an evolutionary algorithm to search the space for diverse generators. On a similar note, Drageset et al~\cite{drageset2019optimising} defined a meta-generator space where each generator is defined as a set of parameters for a constructive level generator for general video game framework~\cite{perez2019general}. They used an optimization algorithm to search the generator space by sampling different levels from each generator. In the end, the algorithm returns the best found level from all the evolved generators.

Cellular Automata can be considered as level generators as they can modify the input to an new output that follow certain rules. These rules can be designed in a way to generate organic like levels~\cite{johnson2010cellular}.  Ashlock~\cite{ashlock2015evolvable} evolved cellular automata rules to find different generators that generate black and white maps with full connectivity. Similarly, Pech et al~\cite{pech2015evolving} and Adams and Louis~\cite{adams2017procedural} generate cellular automata rules to generate mazes with certain features.

We can also see Procedural Content Generation through Machine Learning (PCGML)~\cite{summerville2018procedural} as Procedural Procedural Level Generator Generation. PCGML uses machine learning techniques to train models that can efficiently sample from the learned distribution the content it was trained on. Many different methods were used to generate game levels such as Generative Adverisal Networks~\cite{volz2018evolving}, AutoEncoder~\cite{jain2016autoencoders}, MarkovChains~\cite{snodgrass2014experiments}, LSTM~\cite{summerville2016super}, Bayesian Networks~\cite{guzdial2016game}, etc.

Another way is to represent the generator in form of neural network. Earle~\cite{earle2019using} trained fractal neural networks using A2C~\cite{mnih2016asynchronous} to play SimCity (Will Wright, 1989). This might not look like level generation but if you look at SimCity as city planning problem, then the agent is generating cities. Khalifa et al.~\cite{khalifa2020pcgrl} explored another neural network level generator space similar to Earle's work. They represented level generation as an iterative process where at each step the agent is taking an action to improve the overall level. They used reinforcement learning to train the neural network on 3 different problems with different reward function. The trained networks were able to do learn how to modify the map from random initialization state to playable level in all the 3 problems.

\subsection{Marahel Framework}
Marahel~\cite{khalifa2017marahel} is a constructive level generator description language\footnote{https://github.com/amidos2006/marahel}. Each Marahel script constitutes a level generator. The script defines the generator from a bottom up approach, instead of identifying the requirement of the content, you specify the steps to reach that goal. The language was introduced to help unify the different constructive technique approaches the game designer and developer uses~\cite{short2017procedural} during game development. Marahel levels are expressed as 2D matrix of integers where each integer reflect a certain game entity that is defined in the Marahel script. 

A Marahel script consists of 5 different parts (Metadata, Entities, Regions, Neighborhoods, and Explorers) which will be explained later. Marahel starts by reading the whole script and then creates a $NxM$ 2D matrix of ``undefined'' entities such that $N$ and $M$ are defined in the Metadata. Then, it divides the map into several regions using the information in the Regions section. Finally, Marahel applies the explorers one by one where each explore modifies the 2D matrix based on the provided rules in each explorer.
\subsubsection{Metadata:} contains the allowed sizes of the generated maps.
\subsubsection{Entities:} is a list of different game entities that can be placed in the generated level.
\subsubsection{Regions:} divides the full map using an algorithm (such Binary Space Partitioning~\cite{shaker2016procedural}) into several areas where each area is a group of tiles that can be addressed by itself in the explorers later.
\subsubsection{Neighborhoods:} is a list of relative locations that the explorers can use during generating the map. Relative locations can be used to check certain areas around a certain tile. This is similar to the Cellular Automata neighborhoods used in cave generation~\cite{johnson2010cellular} or convolution filters in Convolutional Neural Networks~\cite{lecun1995convolutional}.
\subsubsection{Explorers:} is the core part of the generation. Explorers visits different tiles in the map/region in a certain order where each visited tile can be modified using a set of input rules. The order of the visited tiles can be defined using some keywords such as ``horizontal'' where it visits all the tiles one by one like scan-lines or ``random'' where it visits tiles in a random order. At each visited tiles, the explorer go over all the rules one by one and stop whenever a rule is satisfied. The rules consists of two parts conditions and executers. Conditions check certain constrains at the visited tiles. For example, ``self(empty)'' checks if the current tile (``self'' neighborhood) is of entity type ``empty''. Executers specify what change should happen to the map relative to that location if the condition is satisfied. For example, ``all(solid)'' will modify a 3x3 grid (``all'' neighborhood) around the current location (including the current location) to be all ``solid'' entity.

\section{Methods}\label{sec:methods}

\begin{figure*}
    \centering
    \begin{subfigure}[t]{.3\linewidth}
        \centering
        \includegraphics[width=\linewidth]{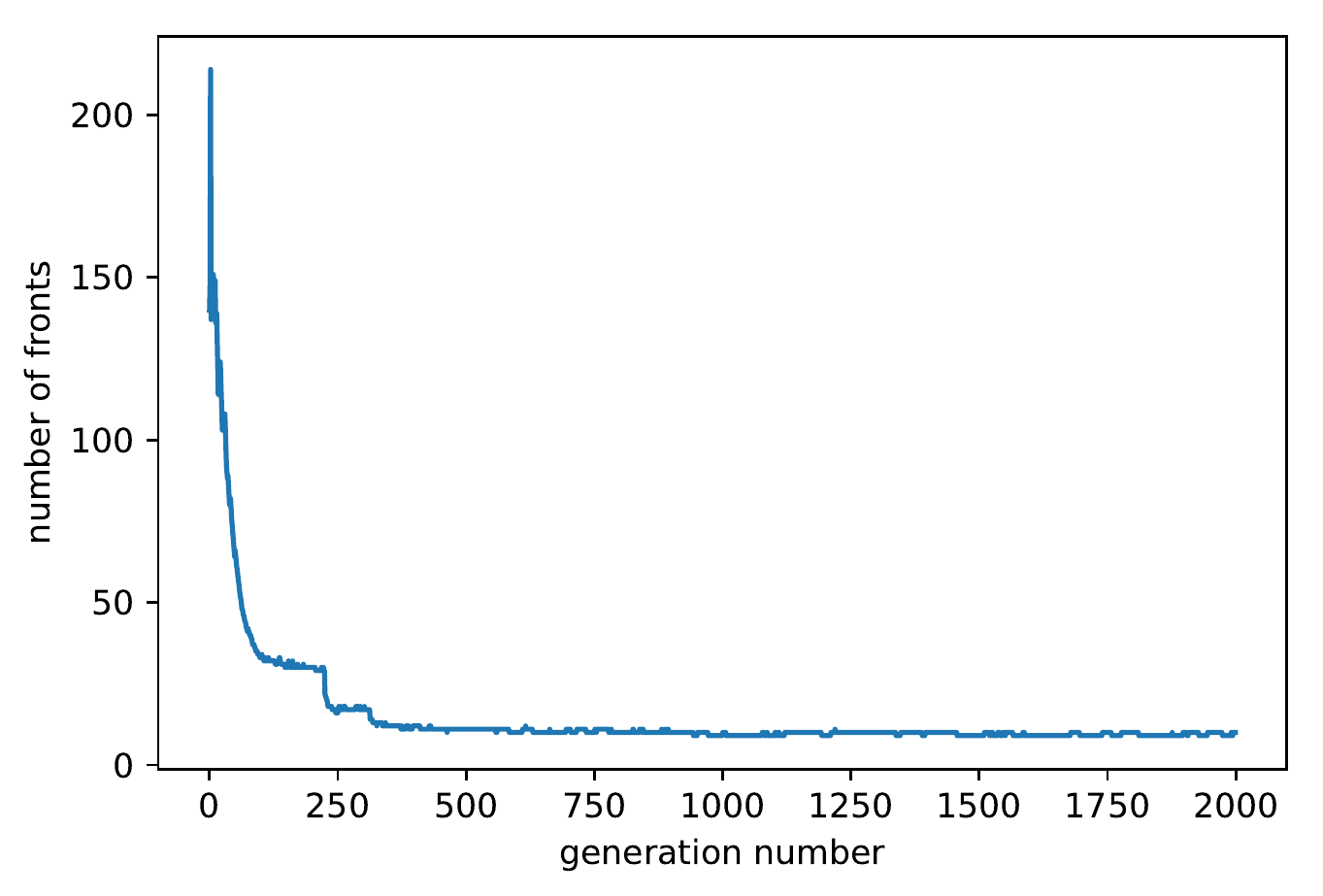}
        \caption{Binary}
        \label{fig:paretoBinary}
    \end{subfigure}
    \begin{subfigure}[t]{.3\linewidth}
        \centering
        \includegraphics[width=\linewidth]{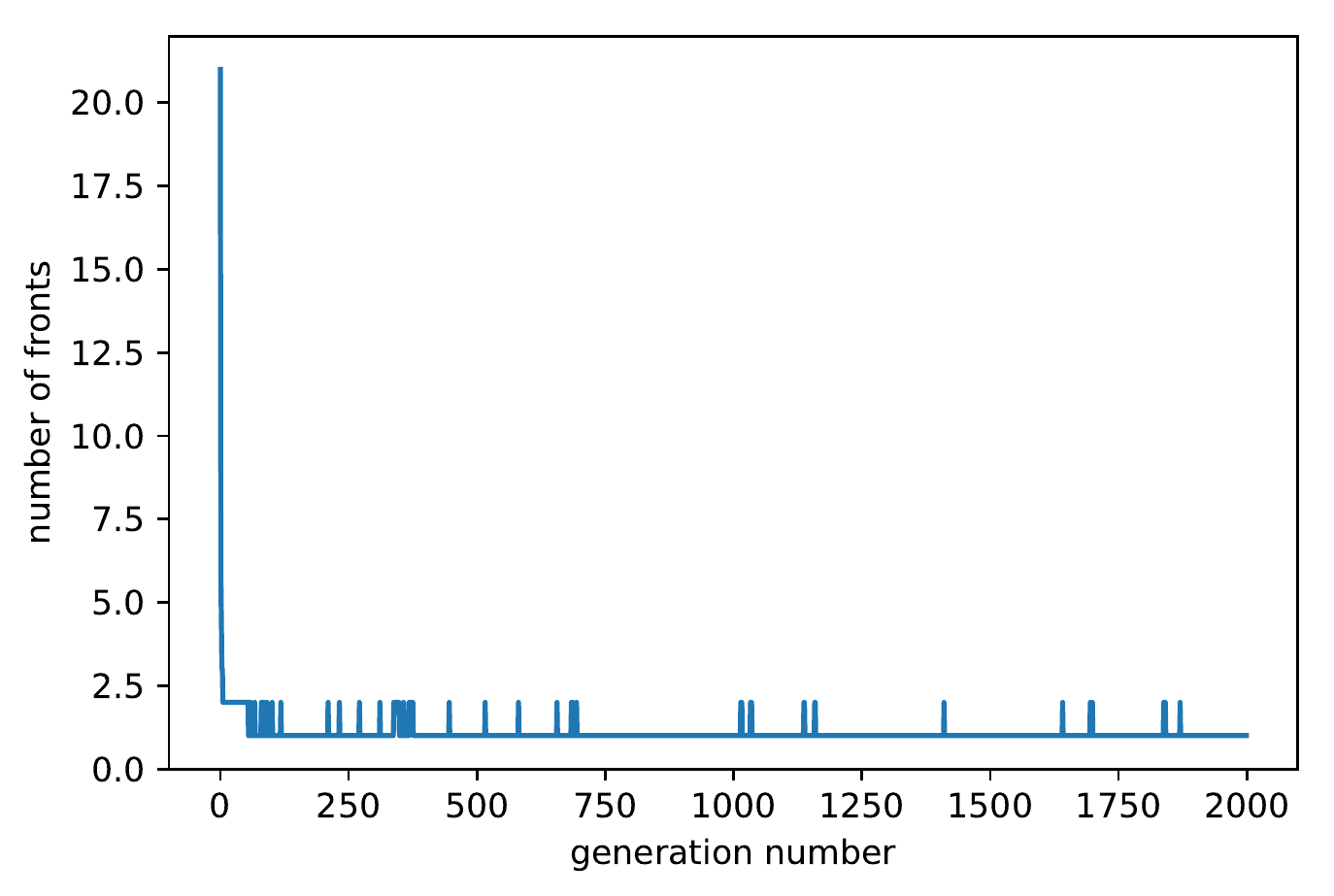}
        \caption{Zelda}
        \label{fig:paretoZelda}
    \end{subfigure}
    \begin{subfigure}[t]{.3\linewidth}
        \centering
        \includegraphics[width=\linewidth]{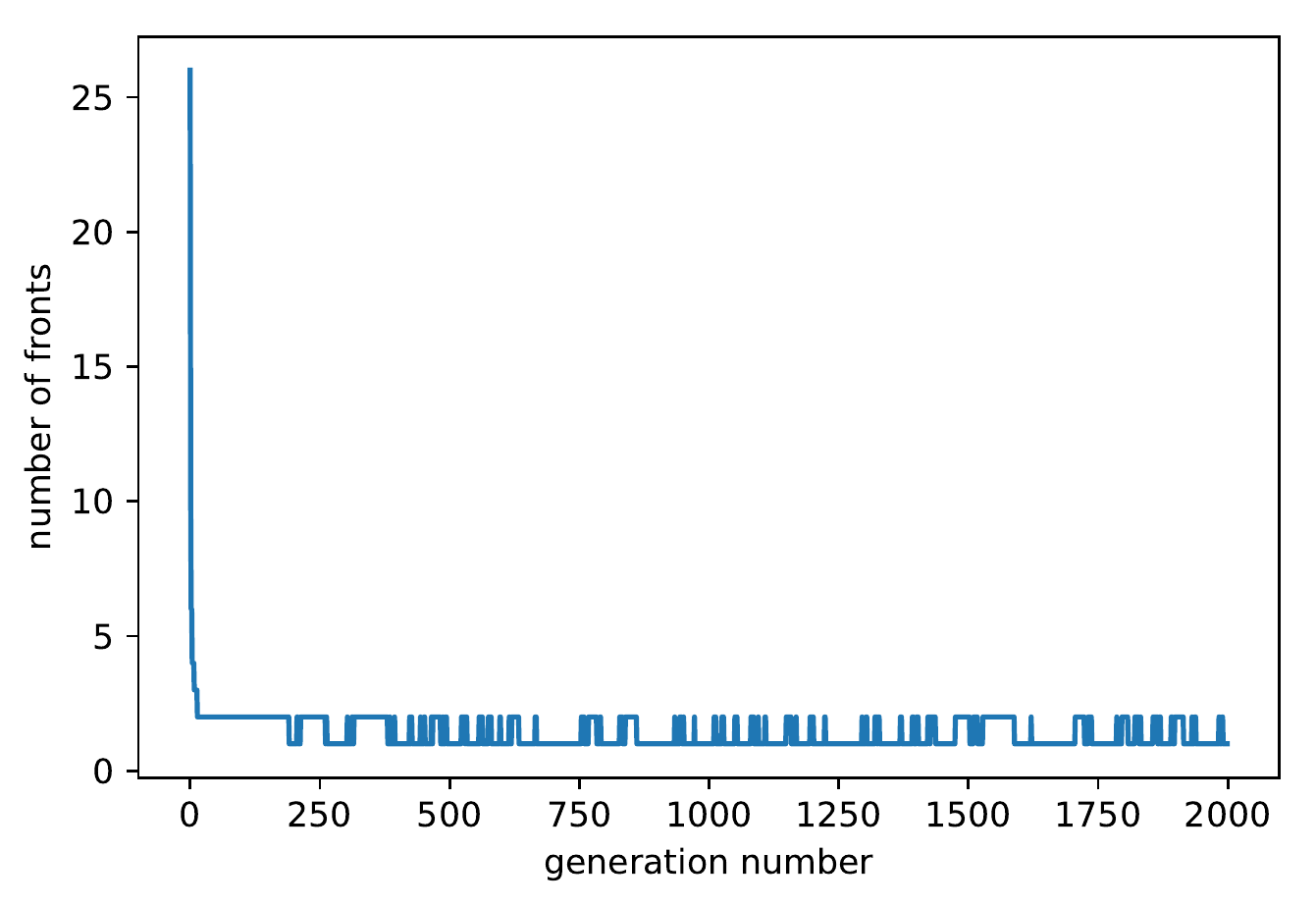}
        \caption{Sokoban}
        \label{fig:paretoSokoban}
    \end{subfigure}
    \caption{Number of Pareto fronts at each generation for all the three problems.}
    \label{fig:numParetoFronts}
\end{figure*}

We use Grammatical Evolution~\cite{ryan1998grammatical,o2001grammatical} to evolve our Marahel scripts. We restricted our evolution to only evolve five different explorers. The reason is to force the evolution to find interesting small scripts than allowing for big ones. We also introduce an explorer before these five to initialize the map with random tiles instead of starting from ``undefined'' map. This step allows the evolution to focus on achieving the target results instead of making sure that the final output doesn't have ``undefined'' entities. There is no regions, so all the explorers are applied on the full map. For neighborhoods, we fixed them to a predefined set of 18 different ones. These neighborhoods covers different local configurations that can be used such as Moore neighborhood, Von Neumann neighborhood, diagonal neighborhood, etc. These neighborhoods are defined using 3 different sizes 1x1, 3x3, and 5x5.

We are going to search for generators for the same three problems introduced in the PCGRL Framework~\cite{khalifa2020pcgrl}. The goal of generation is to find a playable level.
\begin{itemize}
    \item \textbf{Binary:} is a 2D maze and it is the simplest problem. A good level is a level where all the empty tiles are connected using using only cardinal directions (up/down/left/right) and the length longest shortest path in that maze increased by $X$ tiles from the starting state. For example: if we have random map with longest shortest path equal to 10, then the generator has to modify that initial map such that the longest shortest path is equal to 10 +$X$ where $X$ is any positive integer value.
    \item \textbf{Zelda:} is a GVGAI~\cite{perez2019general} port of the dungeon system of The Legend of Zelda (Nintendo, 1986). The goal of the game is to get a key and get to the door without dying from the moving monsters. A good level is a level where there is one player, one key, one door, and the path length between the player and the key plus the path length between the key and the door is at least $X$ steps.
    \item \textbf{Sokoban:} is a port of a Japanese puzzle game by the same name. The goal of the game is to push every crate onto a target location. A good level is a level where there is one player, a number of crates equal to number of targets, and it can be solved with minimum number of steps equal to $X$.
\end{itemize}

The initialization explorer that we added is adjusted similarly to the one used in the PCGRL framework where it is biased to have a good starting state. In Binary, the empty is equal to solid equal to 50\%. In Zelda, Empty is 50\%, Solid is 25\%, Enemies is 10\%, Player is 5\%, Key is 5\%, and Door is 5\%. Lastly in Sokoban, Solid is 40\%, Empty is 45\%, Player is 5\%, Crate is 5\%, and Target is 5\%. The reason is to have same starting point similar to the PCGRL framework which allows us to compare our results. Also, these probabilities generates levels that requires small amount of changes to make it playable therefore helping the evolution.

We decided on using multi-objective Evolution instead of using normal single-objective evolution such as a GA as we found from preliminary experiments that GA doesn't improve much in all the different requirements. We think the reason is that our constrained search space makes it hard to optimize all these values at the same time. An increase in one value will cause a decrease in another one which can be seen in later in section~\ref{sec:results}.

\subsection{Representation}
The chromosome consists of 102 integer numbers; each number has a value between 0 and 49. The first number identify the number of explorers used with maximum of five, the second number is the seed for random number generator, and each 20 integers after that correspond to an explorer. The transformation from integers to the corresponding Marahel script is done using Tracery~\cite{compton2015tracery}. We defined the explorer part as a context free grammar and feed this grammar to the Tracery engine, then we use each integer in the array to guide each non-terminal expansion till all the non-terminal symbols are expanded.

\subsection{Genetic Operators}
We are using two genetic operators: Crossover and Mutation. The crossover operator allows for bigger yet meaningful changes. It can swap either the seed number, number of explorers, or one of the explorers (all the 20 numbers). On the other hand, the mutation operator performs a very small change. It picks a random location from the array and replace it with another random value.

\subsection{Fitness Functions}
In this work, we want to find generators that can produce playable levels for all the three problems. The problem of using solely playability as our fitness is its rough fitness landscape. All the randomly initialized generators for most of the problems will produce 100\% unplayable levels. Having additional fitness functions allow the space to be smoother or optimized toward these ones till reach the goal.

All our fitness function are designed to reflect how close the actual value to the desired value. For example: if we want to have one player, so our fitness function will be 1 if the number of player is 1 and less than one otherwise. The value is calculate using the following equation.
\begin{equation}\label{eq:linearFitness}
    f(x) =
    \begin{cases}
        \frac{range_{min} - x}{range_{min}} & \text{if $x < range_{min}$} \\
        1 & \text{if $range_{min} \leq x \geq range_{max}$} \\
        \frac{x - range_{max}}{max - range_{max}} & \text{if $x > range_{max}$}
    \end{cases}
\end{equation}
where $x$ is the input value to be scaled, $range_{min}$ is the minimal acceptable value, $range_{max}$ is the maximum acceptable value, and $max$ is the maximum possible value. The $f(x)$ is clamped to be always between 0 and 1.

In this work, we evaluate generators instead of single levels, so we need to scale equation~\ref{eq:linearFitness} to work on multiple levels. We solved that by sampling a batch of levels from the generator needed to be evaluated. We calculate the fitness values using equation~\ref{eq:linearFitness} for each level. Finally, we combine these values by taking the average over the sampled levels.

\subsubsection{Binary}
\begin{itemize}
    \item \textbf{Number of Regions:} the number of regions in the generated map. The goal is to have one region so $range_{min}$ equals to $range_{max}$ equals to 1 and $max$ is 10.
    \item \textbf{Path Length Improvement:} the amount of increase in the shortest longest path after the random initialization explorer. The goal is to have an increase of at least 20. To achieve that, $range_{min}$ is equal to 20 and $range_{max}$ is infinity.
\end{itemize}
\subsubsection{Zelda}
\begin{itemize}
    \item \textbf{Number of Players:} the number of players in the generated map. The goal is to have one player. To achieve that, $range_{min}$ equal to $range_{max}$ equal to 1, and $max$ is equal to $10$.
    \item \textbf{Number of Keys:} the number of keys in the generated map. Similar to the number of players, the goal is to have one key.
    \item \textbf{Number of Doors:} the number of doors in the generated map. Similar to the number of players, the goal is to have one door.
    \item \textbf{Number of Enemies:} the number of keys in the generated map. The goal is to have not many enemies and not too few enemies. To achieve that, the $range_{min}$ is 2, $range_{max}$ is 4, and $max$ is equal to 10.
    \item \textbf{Solution Length:} the number of steps the player needs to reach the key then the door. The goal is to have at least 20 steps to finish the level. Similar to path length improvement, we set $range_{min}$ to 20 and $range_{max}$ to infinity.
\end{itemize}
\subsubsection{Sokoban}
\begin{itemize}
    \item \textbf{Number of Players:} the number of players in the generated map. The goal is to have one player. To achieve that, $range_{min}$ equal to $range_{max}$ equal to 1, and $max$ is equal to $10$.
    \item \textbf{Number of Crates:} the number of crates in the generated map. The goal is to have not too many crates and not too few so we set $range_{min}$ to 2, $range_{max}$ to 4, and $max$ to 10.
    \item \textbf{Absolute Difference:} the absolute difference between the number of crates and targets in the generated map. The goal is to have number of crates equal to number of target so the level can be won. To achieve that, we set $range_{min}$ and $range_{max}$ to 0, and $max$ to 10.
    \item \textbf{Solution Length:} the number of steps the player need to win a Sokoban level (all crates are on targets). The goal is to have at least 20 steps to finish the level. Similar to path length improvement, we set $range_{min}$ to 20 and $range_{max}$ to infinity.
\end{itemize}

\section{Results}\label{sec:results}

For the evolution, we used the NSGA-II algorithm~\cite{deb2002nsga}. We used tournament selection of size 2, population size of 500, number of generation equal to 2000, crossover rate equal to 70\%, and mutation rate equal to 30\%. For each problem, we have different size map similar to the same sizes from PCGRL framework. For Binary, the map size is 14x14, while Zelda is 11x7, and finally Sokoban is 5x5. The fitness value is calculate by averaging the fitness over 50 sampled maps from the evaluated generator.

Figure~\ref{fig:numParetoFronts} shows the number of Pareto fronts at each generation. At generation 2000, we can see that for Zelda and Sokoban there is only one front where all the 500 chromosomes are in it. This suggests that there might be more interesting generators that the generator was not able to find. On the other hand, the binary problem ends with having 10 fronts where the first front have 36 chromosomes. We think the reason for that is the number of fitness functions used for Zelda and Sokoban compared to Binary.

\subsection{Binary}

\begin{figure}
    \centering
    \includegraphics[width=.6\linewidth]{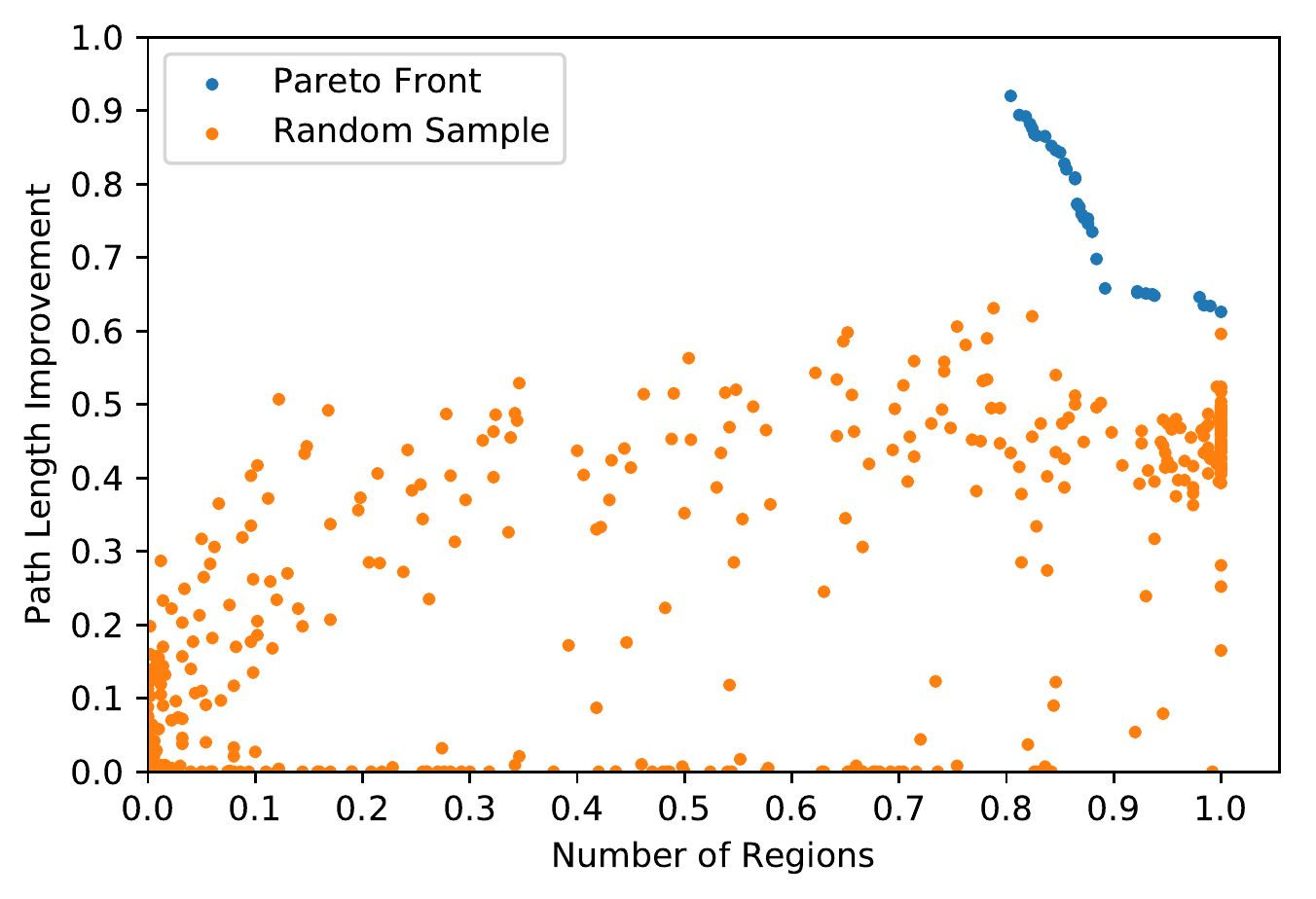}
    \caption{The Pareto front for the Binary Problem for both fitness functions.}
    \label{fig:binaryFront}
\end{figure}

As discussed before in section~\ref{sec:methods}, the current representation and restrictions don't allow us to find a generator that satisfies both fitness functions. Having a high path length leads to having more than one region, while having one region leads to having less than 20 path length improvement. Figure~\ref{fig:binaryFront} shows the Pareto front of the Binary problem after 2000 generation. The Pareto front contains 36 chromosomes out of the 500 while the rest are distributed on the other 8 fronts. 

The figure also shows 500 random sampled generators. We can notice that achieving connectivity (number of regions is equal to 1) is not a hard constraint to satisfy as 11\% (59 out of 500) of the random sampled generators can achieve that. On the other hand, having long path length with connectivity is not an easy task with a maximum of $0.6$ for random sampled generators.

\begin{figure*}
    \centering
    \resizebox{.3\linewidth}{!}{
    \begin{subfigure}[b]{.45\linewidth}
            \centering
\begin{lstlisting}[language=HTML,numbers=none,showstringspaces=false]
{"explorers": [
{
	"type": "connect",
	"parameters": {
	    "repeats": "1",
		"replace": "buffer",
		"directions": "plus",
		"entities": "empty"
	},
	"rules": [
		"self(any) -> vert(empty)"
	]
},
{
	"type": "connect",
	"parameters": {
	    "repeats": "1",
	    "replace": "same",
		"directions": "plus",
		"entities": "empty"
	},
	"rules": [
		"self(out) -> plusnc(solid|empty)"
	]
},
{
	"type": "horz",
	"parameters": {
	    "repeats": "1",
		"replace": "buffer"
	},
	"rules": [
		"plusfive(empty) -> down(solid)"
	]
}]}
\end{lstlisting}
    \end{subfigure}}
    \begin{subfigure}[b]{.3\linewidth}
        \centering
        \includegraphics[width=.45\linewidth]{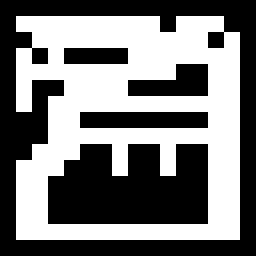}
        \includegraphics[width=.45\linewidth]{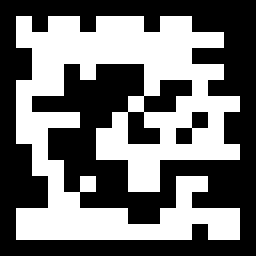}
        \includegraphics[width=.45\linewidth]{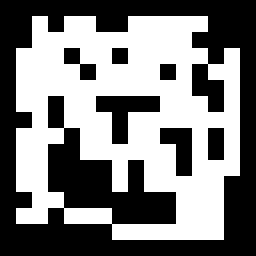}
        \includegraphics[width=.45\linewidth]{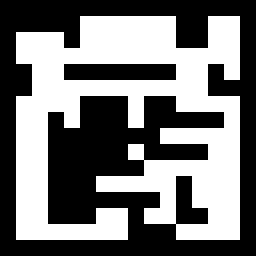}
        \includegraphics[width=.45\linewidth]{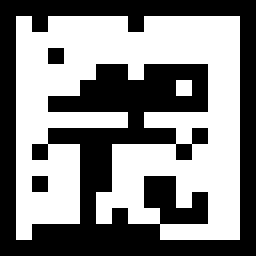}
        \includegraphics[width=.45\linewidth]{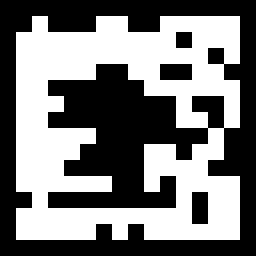}
        \includegraphics[width=.45\linewidth]{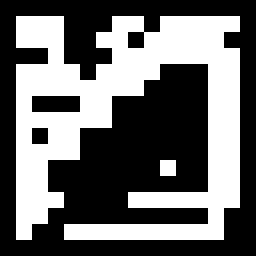}
        \includegraphics[width=.45\linewidth]{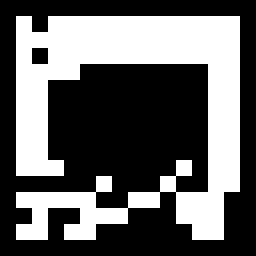}
    \end{subfigure}
    \caption{The evolved binary generator and several different generated examples. On the left, the evolved explorers are shown. On the right, several different examples produced by the generator. The black tiles are solid, while the white tiles are empty. This generator has number of regions fitness value equal to 0.8 and path length improvement fitness value equal to 0.92}
    \label{fig:binaryExamples}
\end{figure*}

Figure~\ref{fig:binaryExamples} shows one of the Pareto front generators that have the highest path length improvement 0.92 (an average of 18 increase) and connectivity of 0.8 (an average of 2 regions). Looking at the generator, the generator has 3 explorers. The first explorer connect the empty tiles of the random initialized level using 3 empty tiles in a form of a vertical line. This explorer leads to a fully connected level with big open space as the connection uses vertical lines instead of just placing a single tile. The second explorer is another connecting one but since the first one connected everything then it won't be executed. Finally, the last explorer goes on every tile in the map and check if it is empty and surrounded by empty tiles using 5x5 Moore neighborhood, it will convert the tile below the current location to solid. This last explorer is the reason for having more than 1 region as it could isolate an area by accident but at the same time, it is what guarantees the long path as it doesn't allow big open areas. Looking at the examples in figure~\ref{fig:binaryExamples}, we can see that the long vertical connections leaded to having a circular layouts and we can see a small few disconnected areas due to the last explorer.

\subsection{Zelda}

\begin{figure*}
    \centering
    \begin{subfigure}[b]{.3\linewidth}
        \centering
        \includegraphics[width=\linewidth]{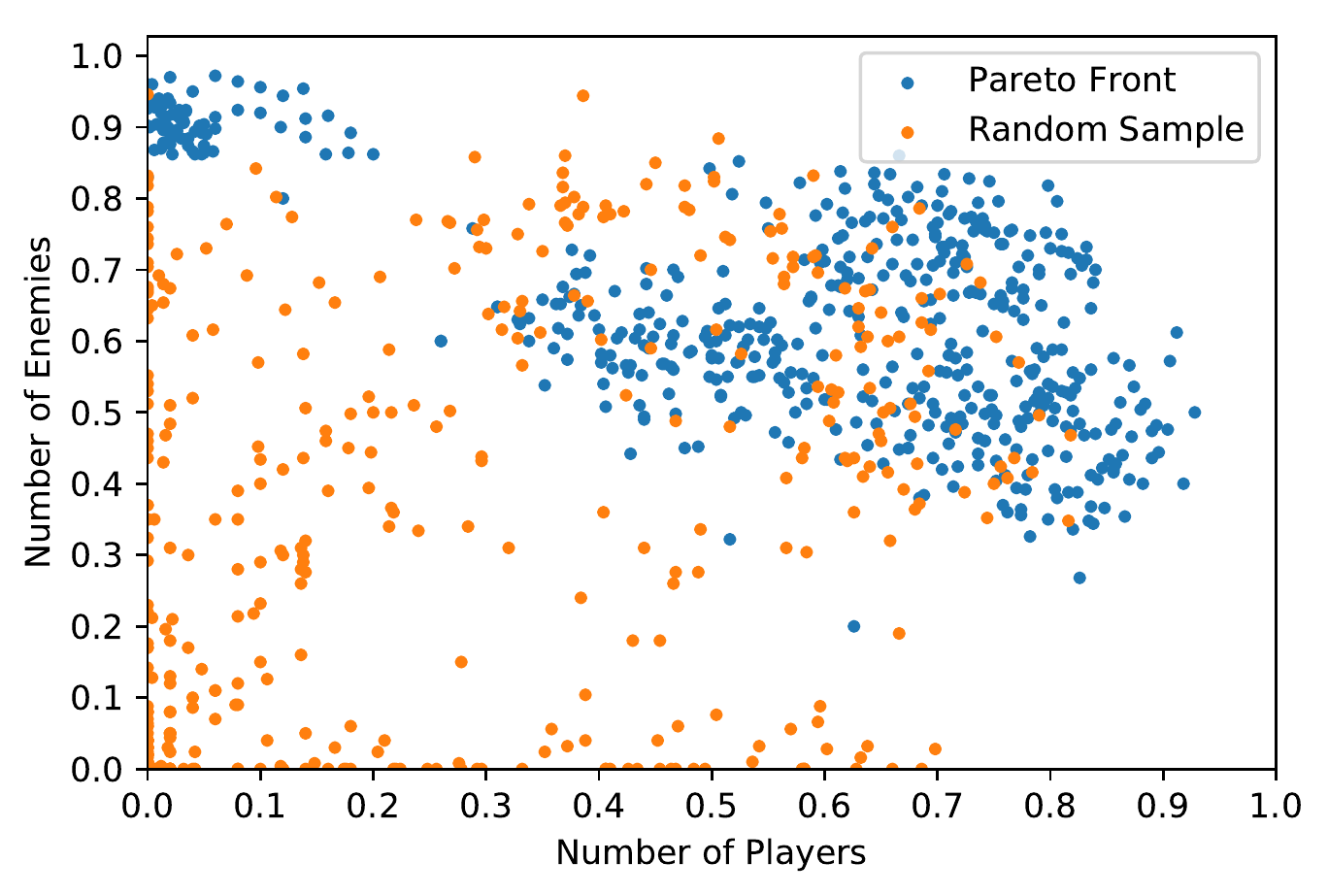}
        \caption{}
        \label{fig:zeldaFront03}
    \end{subfigure}
    \begin{subfigure}[b]{.3\linewidth}
        \centering
        \includegraphics[width=\linewidth]{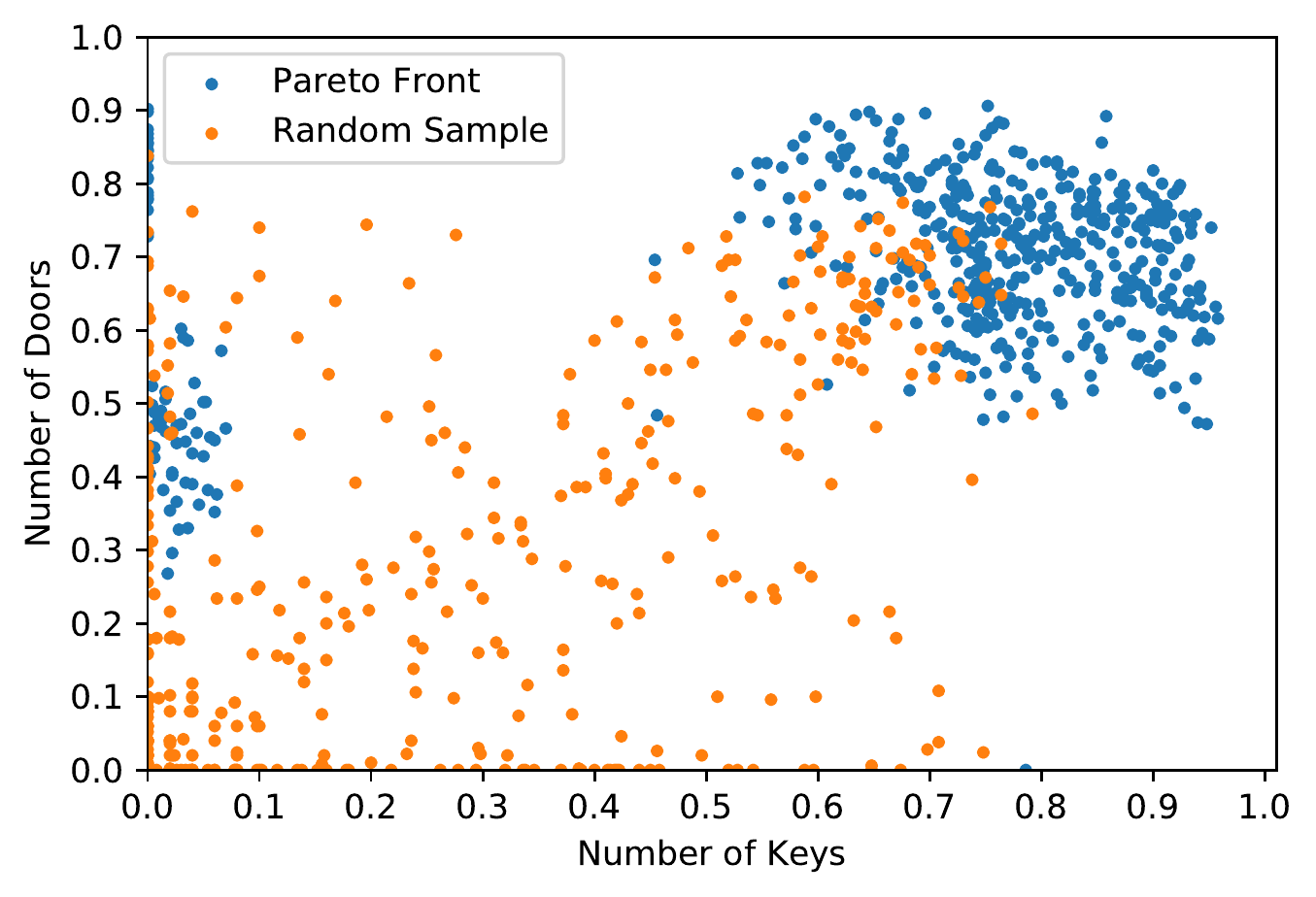}
        \caption{}
        \label{fig:zeldaFront12}
    \end{subfigure}
    \begin{subfigure}[b]{.3\linewidth}
        \centering
        \includegraphics[width=\linewidth]{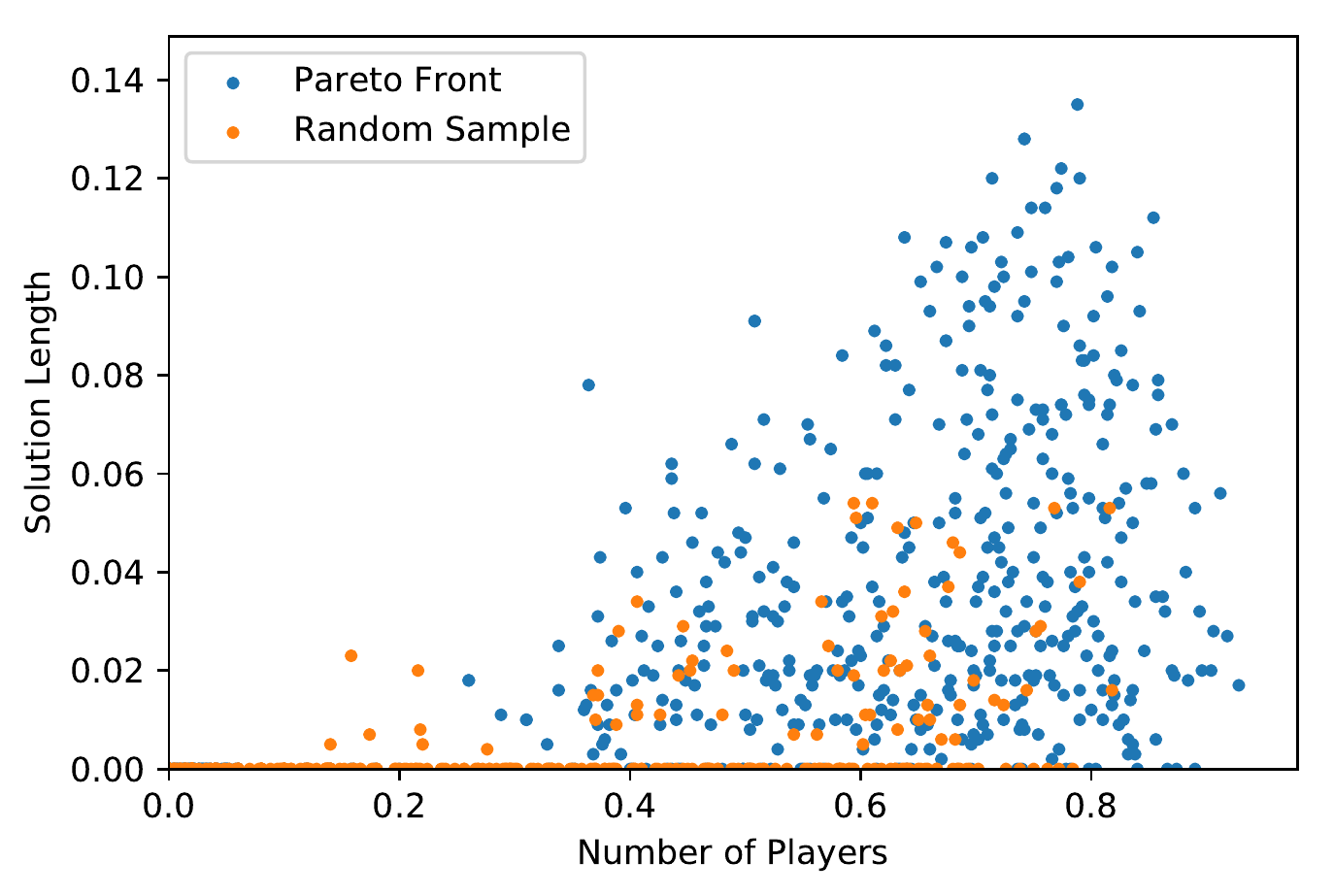}
        \caption{}
        \label{fig:zeldaFront04}
    \end{subfigure}
    \centering
    \caption{The Pareto front for the Zelda Problem for different combination of the fitness functions.}
    \label{fig:zeldaFronts}
\end{figure*}

For the Zelda problem, at the last generation all the chromosomes (500) exists in one front which reflects that there is more chromosomes that can be explored. We decided to show all the Pareto Front for the couple of the fitness function combinations. Figure~\ref{fig:zeldaFront03} shows the Pareto front between number of player and number of enemies fitness functions. It is interesting to see that it is inversely proportional as it means that having more enemies means less chance of having a single player. This makes sense as having more enemies means less tiles for the player avatar.

Figure~\ref{fig:zeldaFront12} shows the Pareto front between number of doors and number of keys fitness function. Interesting enough, the graph is directly proportional everywhere except at the beginning where there is a group of generators that have high fitness for the number of doors but zero fitness for the number of keys. These could be appearing due to other non dominated dimensions such as number of enemies which might be overwriting the keys in the levels.

Figure~\ref{fig:zeldaFront04} shows the Pareto front between the number of players and solution length. The solution length fitness is pretty low overall with maximum of $0.14$. This doesn't mean it is unplayable, it also could mean very short length. Looking at the figure, it is obvious when we have a low number of player fitness, we also have a low solution length as you can't play a level if you don't have one player. On the other hand, with high number of players, the solution length is bouncing between almost 0 and 0.14. This noise is due to the other two fitness function: number of keys and number of doors as the only way to have a solution length is to have one player, one key, and one door. This makes it a very hard fitness function to satisfy causing these low fitness values.

Looking at the 500 random sampled generator, we can notice that it can cover huge parts of the space but we think it is not able to satisfy all of these constraints at the same time because solution length fitness for all these ones doesn't increase than $0.06$.

\begin{figure*}
    \centering
    \resizebox{.38\linewidth}{!}{
    \begin{subfigure}[b]{.45\linewidth}
            \centering
\begin{lstlisting}[language=HTML,numbers=none,showstringspaces=false]
{"explorers": [
{
	"type": "greedy",
	"parameters": {
		"repeats": "3",
		"replace": "buffer",
		"directions": "all",
		"heuristics": "dist(empty)"
	},
	"rules": [
		"diagfive(key) -> down(door)"
	]
},
{
	"type": "connect",
	"parameters": {
		"repeats": "1",
		"replace": "buffer",
		"directions": "all",
		"entities": "empty"
	},
	"rules": [
		"noise>=-0.5  -> left(solid)"
	]
}]}
\end{lstlisting}
    \end{subfigure}}
    \begin{subfigure}[b]{.36\linewidth}
        \centering
        \includegraphics[width=.4\linewidth]{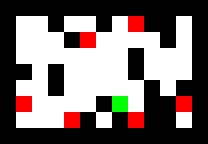}
        \includegraphics[width=.4\linewidth]{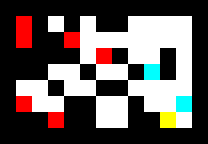}
        \includegraphics[width=.4\linewidth]{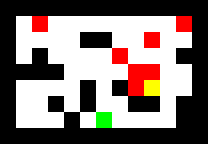}
        \includegraphics[width=.4\linewidth]{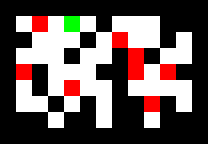}
        \includegraphics[width=.4\linewidth]{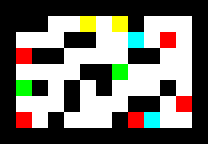}
        \includegraphics[width=.4\linewidth]{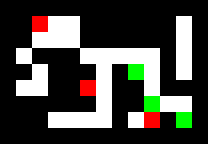}
        \includegraphics[width=.4\linewidth]{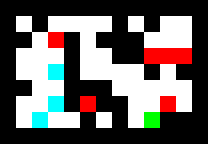}
        \includegraphics[width=.4\linewidth]{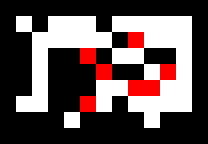}
        \includegraphics[width=.4\linewidth]{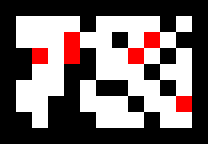}
        \includegraphics[width=.4\linewidth]{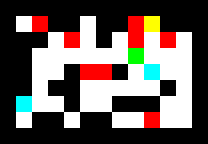}
    \end{subfigure}
    \caption{The evolved zelda generator and several different generated examples. On the left, the evolved explorers are shown. On the right, several different examples produced by the generator. Black tiles are solid, white tiles are empty, green tiles are player, red tiles are enemies, yellow are keys, and cyan are doors. This generator has number of players fitness value equals to 0.79, number of keys fitness value equals to 0.67, number of doors fitness value equals to 0.7, number of enemies fitness value equals to 0.59, and solution length fitness value equal to 0.14}
    \label{fig:zeldaExamples}
\end{figure*}

Figure~\ref{fig:zeldaExamples} shows a Marahel Zelda generator evolved after 2000 generation. We picked this generator as it has the highest solution length fitness ($0.14$). Looking on the 10 generated examples, we can notice that non of them are playable but some can be easily fixed to be playable. The interesting thing about this generator example is it have small number of players and keys and doors which make it have a high chance to generate playable levels. Looking into the generator itself, we can see it is more of an eraser. It depends on the starting noise and it tries to substitute some of the tiles by solid using a noise function while moving on the path to connect entities which cause it to have high fitness values.

\subsection{Sokoban}

\begin{figure}
    \centering
    \begin{subfigure}[b]{.45\linewidth}
        \centering
        \includegraphics[width=\linewidth]{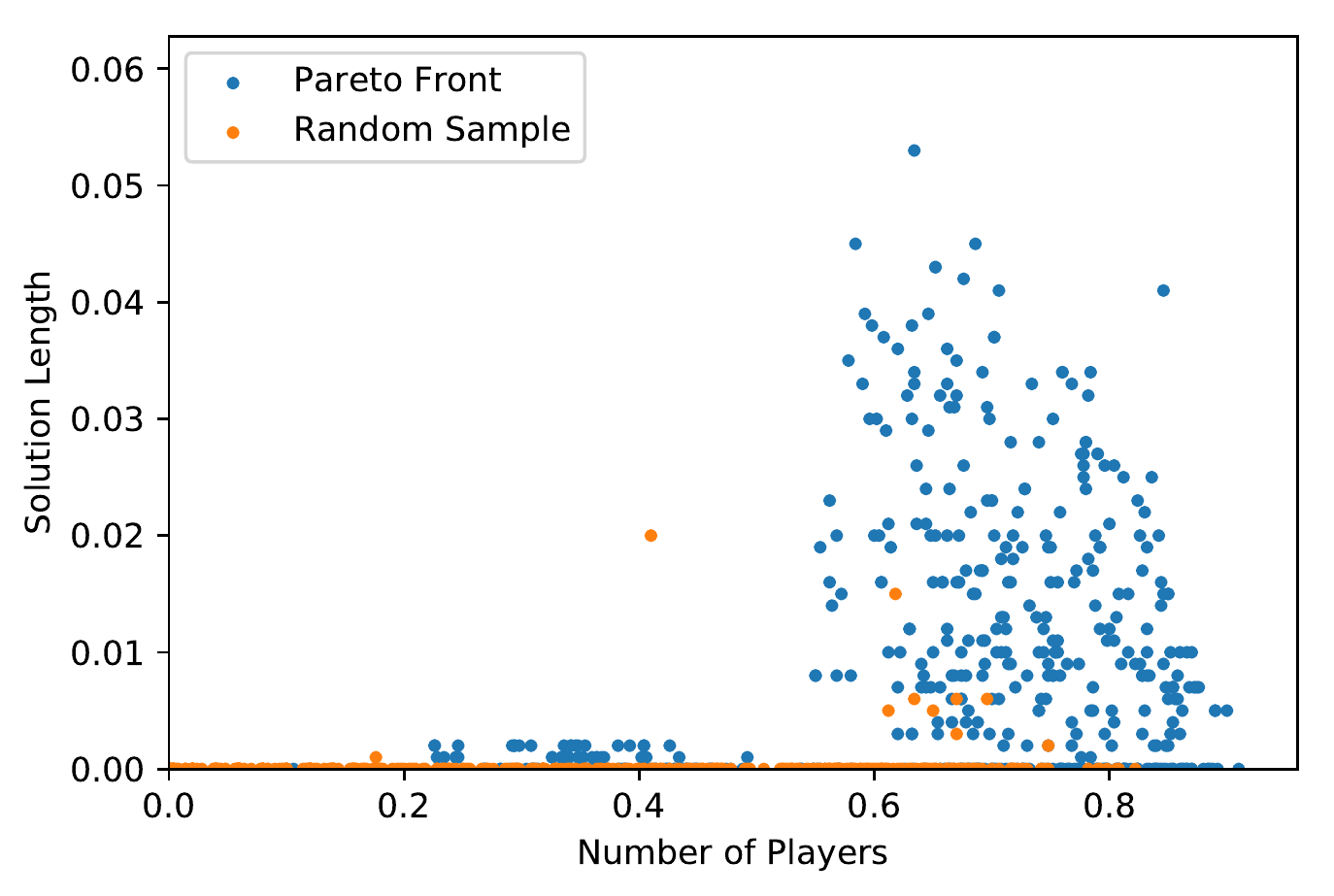}
        \caption{}
        \label{fig:sokobanFront03}
    \end{subfigure}
    \begin{subfigure}[b]{.45\linewidth}
        \centering
        \includegraphics[width=\linewidth]{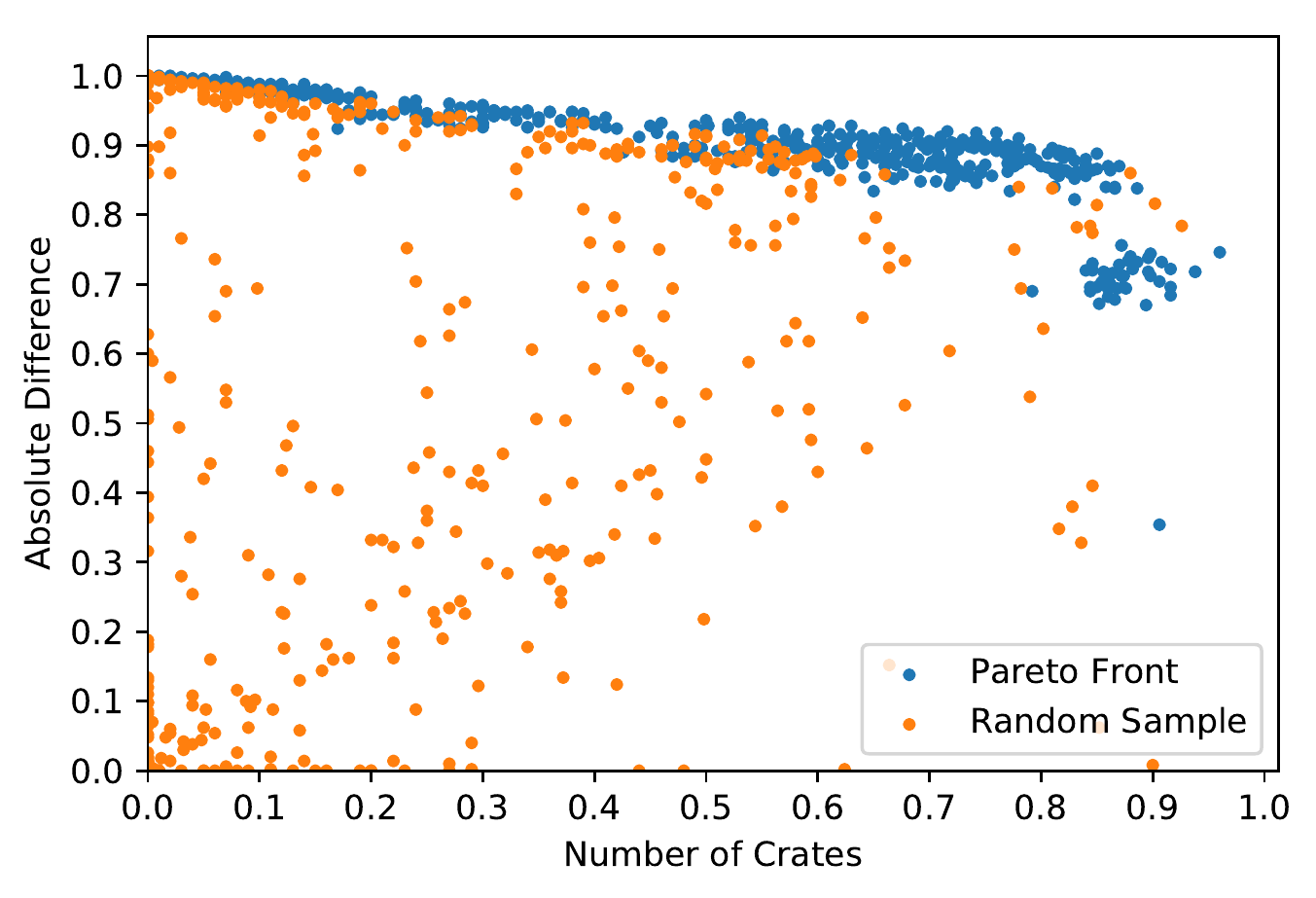}
        \caption{}
        \label{fig:sokobanFront12}
    \end{subfigure}
    \centering
    \caption{The Pareto front for the Sokoban Problem for different combination of the fitness functions.}
    \label{fig:sokobanFronts}
\end{figure}

\begin{figure*}
    \centering
    \resizebox{.3\linewidth}{!}{
    \begin{subfigure}[b]{.45\linewidth}
            \centering
\begin{lstlisting}[language=HTML,numbers=none,showstringspaces=false]
{"explorers":[
{
    "type": "connect",
    "parameters": {
        "repeats": "1",
        "replace": "same",
        "directions": "plus",
        "entities": "empty",
        "out": "crate",
        "changes": "4"
    },
    "rules": [
        "noise>0.3,up(player) -> up(target)",
        "noise<=0.6, random>=0.5,vertl(crate) -> diag(empty)"
    ]
},
{
    "type": "connect",
    "parameters": {
        "repeats": "1",
        "replace": "same",
        "directions": "plus",
        "entities": "empty"
    },
    "rules": [
        "allfive(empty) -> down(solid)"
    ]
}]}
\end{lstlisting}
    \end{subfigure}}
    \begin{subfigure}[b]{.35\linewidth}
        \centering
        \includegraphics[width=.33\linewidth]{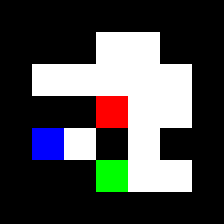}
        \includegraphics[width=.33\linewidth]{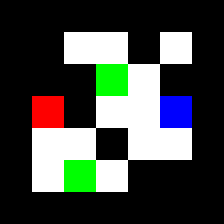}
        \includegraphics[width=.33\linewidth]{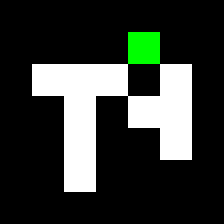}
        \includegraphics[width=.33\linewidth]{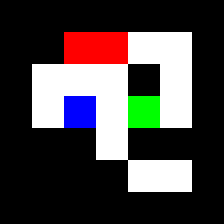}
        \includegraphics[width=.33\linewidth]{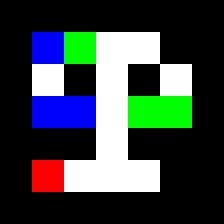}
        \includegraphics[width=.33\linewidth]{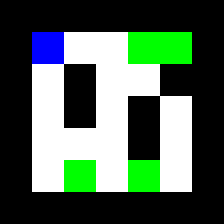}
        \includegraphics[width=.33\linewidth]{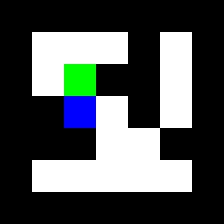}
        \includegraphics[width=.33\linewidth]{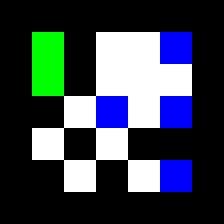}
    \end{subfigure}
    \caption{The evolved Sokoban generator and several different generated examples. On the left, the evolved explorers are shown. On the right, several different examples produced by the generator. Black tiles are solid, white tiles are empty, green tiles are player, red tiles are crates, and blue tiles are targets. This generator has number of players fitness value equals to 0.69, number of crates fitness value equals to 0.66, absolute difference fitness value equal to 0.87, and solution length fitness value equal to 0.045.}
    \label{fig:sokobanExamples}
\end{figure*}

Similar to Zelda, all the 500 chromosomes appear in the first front. Figure~\ref{fig:sokobanFronts} shows the Pareto front for the four different fitness functions. Figure~\ref{fig:sokobanFront03} shows the Pareto front between number of player fitness value and solution length fitness value. It is obvious that having higher number of players will cause the solution length value to increase (you can't play a level if you don't have one player). The solution length value is very noisy and with peak of 4.5\% around 0.6 number of player value. Similar to Zelda, the low solution length fitness value due to the cascaded fitness as a level is only playable if it has one player, number of crates more than 0 and equal to number of targets. Even when all this happens, the level have a higher chance to be unplayable compared to Zelda levels as crates can start in a locked position not allowing them to move even when all the constraints are satisfied. For example: the top left level on in figure~\ref{fig:sokobanExamples} satisfies all the playability constraints but you can't win it as the crate (red tile) can't be moved.

Figure~\ref{fig:sokobanFront12} shows the Pareto front between number of crate fitness value and absolute difference fitness value. The relation is inversely proportional as having higher number of crates fitness value causes the absolute difference value decrease. This is due to having more crates will increase the risk of having more errors in the generated levels (the number of crates are not equal to the number of targets). On the other hand, having a generator that produces no crates and no targets is a very easy task (a generator that erase everything). This generator will always be in the front as it will always have absolute difference fitness value equal to 1 which no other generator achieved it.

Similar to Zelda, the 500 random sample generator can satisfy many of the constraints for playability but is not able to find many playable levels ($0.02$). We think that it was not able to find generators that satisfy all the different constraints at the same time.

Figure~\ref{fig:sokobanExamples} shows the evolved Sokoban generator with 8 different generated levels using that generator. Similarly, we picked this generator as it has the highest solution length fitness value ($0.045$). The generator is a bit simple and it starts by trying to connect between isolated empty areas in the map. The explorer uses a noise function with some constraints on the surrounding tiles to either add empty or target tiles along the connecting path. The second explorer will only work if the first explorer failed to connect all the empty tiles. In that case, the second generator will move along the connecting path and add solid tiles if there is too big of empty space. Similar to Zelda, this generator depends highly on the starting level as most of the conditions are only valid in certain cases and not all the time. For example, if all the empty tiles are connected, neither of these explorers will run.

\section{Conclusion}
This paper introduced a multi-objective optimization method to evolve constructive level generators. The generators used Marahel~\cite{khalifa2017marahel} as their space representation. We restricted the evolution to a small size Marahel scripts to force the evolution to find understandable and efficient generators. The results shows that our restrictions might have caused having a Pareto front and not become able to find a generator that can achieve all the fitness functions 100\%. We also see that only the binary problem was able to explore most of its Pareto front while for Zelda and Sokoban the results are a subset of the full front. The final generator for the Binary problem is interesting as it resulted into these long loopy dungeons. On the other hand, the Zelda generator act as an eraser which erase extra objects from the random initialization, while the Sokoban generator just resampled the level from a different distribution that have higher chance to be playable levels.

The evolution of Marahel resulted into a more understandable generator compared to other techniques~\cite{khalifa2020pcgrl}. The interpretability of Marahel language is a big advantage as we can debug these generators or edit them easily. Our representation and restrictions helped the evolution to find small concise generator that can be understood easily but at the same time it was harder to search the space. %This can be noticed from the fitness functions, they created a Pareto front instead of working in tandem. 
It would be interesting to experiment with less restricted evolution and try to see if this will change the results. We also noticed that having an initialization explorer that initializes the map before the evolved explorers starts, helped to find generators that react to the current initialization by erasing instead of adding (as most of the fitness functions need less number of entities than the starting state). It would be interesting to remove that initialization generator and see if we can achieve different results. For example, generators that try to add more entities than erasing. Another idea, we would like in the future to try to change the fitness function to be more about improvement (similar to path length improvement) instead of optimizing towards a certain value (like number of players, number of regions, etc). We think the improvement fitness functions help the evolution to find more interesting layouts and levels as it doesn't depend on the starting state. One last thing, the average operator (used to aggregate the fitness of the generated sample maps) sometimes biases the generation towards mediocre generators. We think that using different type of operator like mixmin operator (mixing the average value and the minimum value) might forces the generator to move away from these mediocre generators.

\begin{acks}
% The king wants to thank the reviewers for all their great work during the COVID-19 pandemic.
The authors acknowledge the financial support from NSF Award number 1717324 - ``RI: Small: General Intelligence through Algorithm Invention and Selection.''.
\end{acks}
\bibliographystyle{ACM-Reference-Format}
\bibliography{sample-base}

\end{document}